\documentclass[preprint,12pt]{elsarticle}
\usepackage{subfigure}
\usepackage[T1]{fontenc}
\usepackage{amsmath}
\usepackage{float}
\usepackage{graphicx}
\usepackage{multirow}

\usepackage{amssymb}
\usepackage{amsmath}

\journal{Pattern Recognition}

\begin{document}

\begin{frontmatter}



\title{Handwritten Text Recognition for Low Resource Languages}


\author{Sayantan Dey} 

\affiliation{organization={Indian Institute of Technology Roorkee},
            addressline={Haridwar}, 
            city={Roorkee},
            postcode={247667}, 
            state={Uttarakhand},
            country={India}}
\author{Alireza Alaei} 

\affiliation{organization={Faculty of Science and Engineering},
            addressline={Southern Cross University}, 
            city={Gold Coast},
            postcode={4225}, 
            state={QLD},
            country={ Australia}}
\author{Partha Pratim Roy} 

\affiliation{organization={Indian Institute of Technology Roorkee},
            addressline={Haridwar}, 
            city={Roorkee},
            postcode={247667}, 
            state={Uttarakhand},
            country={India}}

\begin{abstract}
Despite considerable progress in handwritten text recognition, paragraph-level handwritten text recognition, especially in low-resource languages, such as Hindi, Urdu and similar scripts, remains a challenging problem. These languages, often lacking comprehensive linguistic resources, require special attention to develop robust systems for accurate optical character recognition (OCR). This paper introduces BharatOCR, a novel segmentation-free paragraph-level handwritten Hindi and Urdu text recognition. We propose a ViT-Transformer Decoder-LM architecture for handwritten text recognition, where a Vision Transformer (ViT) extracts visual features, a Transformer decoder generates text sequences, and a pre-trained language model (LM) refines the output to improve accuracy, fluency, and coherence. Our model utilizes a Data-efficient Image Transformer (DeiT) model proposed for masked image modeling in this research work. In addition, we adopt a RoBERTa architecture optimized for masked language modeling (MLM) to enhance the linguistic comprehension and generative capabilities of the proposed model. The transformer decoder generates text sequences from visual embeddings. This model is designed to iteratively process a paragraph image line by line, called implicit line segmentation. The proposed model was evaluated using our custom dataset ('Parimal Urdu') and ('Parimal Hindi'), introduced in this research work, as well as two public datasets. The proposed model achieved benchmark results in the NUST-UHWR, PUCIT-OUHL, and Parimal-Urdu datasets, achieving character recognition rates of 96.24\%, 92.05\%, and 94.80\%, respectively. The model also provided benchmark results using the Hindi dataset achieving a character recognition rate of 80.64\%. The results obtained from our proposed model indicated that it outperformed several state-of-the-art Urdu text recognition methods.
\end{abstract}




\end{frontmatter}




\section{Introduction}
Since the emergence of digital technology, optical character recognition (OCR) has been the subject of extensive research \cite{ref_article30}. OCR aims to  to transform scanned document images into machine-readable text so that a vast amount of unstructured handwritten or printed text can be analyzed. Historically, handwritten text recognition, including Urdu and Hindi scripts, has been approached through segmentation at incremental levels of granularity—line, word, and character \cite{ref_article11,ref_article19, ref_article25, ref_article26}. Convolutional neural networks (CNNs) with long short-term memory (LSTM) networks were commonly used in the initial wave of deep learning-based OCR models \cite{ref_article31,ref_article32}. Using region proposals for better feature extraction, Region-Based CNNs (RCNNs) \cite{ref_article31} enhanced this architecture. However, as these models were trained character-by-character rather than word-by-word, they had a significant drawback: they ultimately required a separate language model. As a result, Transformer-based OCR (TrOCR) and other transformer-based OCR models were proposed in the literature \cite{ref_article17}. In this approach, a pre-trained language model was employed for text production and substitutes a pre-trained Vision Transformer in the CNN layers. While improving over its predecessor, it has faced the inherent challenge of accurately defining and isolating individual entities (lines, words, and characters) for effective segmentation at each level. In contrast, we propose a new segmentation-free text recognition approach that directly recognizes text from paragraph images without explicit segmentation during the training and decoding phases.

However, the leap to paragraph-level recognition introduces new complexities, such as varying text region/line numbers, diverse layout patterns and skews, and the imperative of defining a coherent reading order. To address these issues, we propose a model that adeptly navigates these challenges using a vision-language model for image-to-text generation, where a pre-trained image encoder extracts visual features using learned tokens. A transformer-based decoder then generates an initial text output that undergoes contextual refinement. Finally, a pre-trained language model is considered to enhance coherence and fluency, ensuring high-fidelity text generation for document understanding and OCR-based tasks. This approach circumvents the error compounding of traditional multi-stage segmentation-recognition approaches. Our approach distinguishes itself by processing entire paragraphs of Handwritten Urdu and Hindi text through a segmentation-free transformer, simplifying the recognition pipeline while enhancing the capability of the model to handle the linguistic and stylistic diversity of both scripts. The complex ligatures and cursive nature of Urdu, alongside the distinct conjunct formations and diacritic variations in Hindi, are effectively addressed within our unified framework. Furthermore, preliminary pre-training on paragraph images is incorporated to elevate recognition accuracy, paving the way for seamless paragraph-level recognition across both languages.

This paper presents a novel approach to handwritten Urdu and Hindi text recognition (HUHTR), moving beyond traditional segmentation-based methodologies in favor of an end-to-end segmentation-free text recognition framework. The contributions of this paper are three-fold:
\begin{enumerate}
 \item \textbf{Introduction of a robust methodology}: This study introduces a novel segmentation-free framework for recognizing handwritten Urdu characters, words, and lines at the paragraph level. 
 \item \textbf{Inclusion of a Large Language Model}: We incorporate the RoBERTa base language model, which has been pre-trained using masked language modeling to improve the comprehension of Urdu and Hindi text, enabling more accurate interpretations of the subtle language features.
  \item \textbf{New dataset created - `Parimal Urdu' and 'Parimal Hindi'}:  Central to our research is the development of a new dataset composed of a diverse array of handwritten Urdu and Hindi pages. This dataset comprises 500 pages contributed by ten individuals from various age groups in each language, ensuring a wide variety of natural handwriting styles. Each contributor was carefully selected to represent different writing variations, making this dataset a valuable resource for handwriting recognition research.
\end{enumerate}

The remainder of the paper is outlined as follows: Section \ref{Related Work} provides an overview of the related work on Urdu text recognition. Section \ref{Architecture} details the description of our proposed model. Section \ref{Experimental Analysis and Discussion} discusses datasets, evaluation metrics, data augmentation, and experimental results and comparative analysis. Section \ref{Conclusion and Future Scope} concludes the paper and presents future research directions.

\section{Related Work}
\label{Related Work}
Due to various complexities and challenges, such as cursive text, complex ligatures, and varying writing styles, UHTR remains a difficult task \cite{ref_article21}. Nevertheless, researchers have explored handwritten Urdu OCR in the literature \cite{ref_article11,ref_article18}. In the early stages, conventional machine learning models, such as SVM \cite{ref_article22}, were used for UHTR. However, a significant shift from traditional methods led to the introduction of holistic approaches utilizing hybrid CNN-RNN networks \cite{ref_article15}. These modern methods employed CNNs to extract essential visual features from input images, which are then processed through Recurrent Neural Networks (RNNs) \cite{ref_article13} to capture contextual information essential for transcription layers. Contemporary state-of-the-art Urdu OCR models incorporated various network architectures for feature extraction and sequential modeling. Bi-directional LSTM (BiLSTM) \cite{ref_article12} networks and Multi-Dimensional LSTM (MDLSTM) \cite{ref_article14} networks were particularly favored for sequential processing, with most models relying on a Connectionist Temporal Classification (CTC) layer for final output transcription.

In contrast, some models explored utilizing DenseNet \cite{ref_article11} and GRU \cite{ref_article16} networks, coupled with an attention-based decoding layer, marking a shift towards integrating attention mechanisms \cite{ref_article11} to improve transcription accuracy. This trend of employing various deep learning architectures \cite{ref_article10,ref_article9,ref_article8} mirrored the development trajectory observed in OCR technologies for languages with scripts similar to Urdu, such as Arabic. Moreover, to address the scarcity of extensive labeled datasets in this domain, enhanced prototypical network architectures were explored for few-shot recognition of handwritten Urdu characters, aiming to improve the classification performance with limited samples \cite{ref_article34}.  However, despite these advances, adapting Arabic OCR techniques to Urdu text often resulted in lower accuracy levels, emphasizing the unique challenges posed by Urdu script that were not fully addressed by existing methods. In addition, although each proposed model aspired to enhance Urdu OCR, a key limitation is the reliance on approaches designed for other languages without adequately tailoring them to accommodate the challenges inherent in Urdu script. 

Research in handwritten Hindi character recognition has seen significant advancements through various approaches. Initial research focused on feature extraction techniques combined with conventional classifiers. For instance, studies employed k-means clustering and structural features to recognize handwritten Hindi characters \cite{ref_article35}. Reddy and Babu \cite{ref_article28} developed a handwritten Hindi character recognition system utilizing CNNs optimized with RMSprop and Adam, achieving high accuracy. In addition, Sharma and Ramakrishnan \cite{ref_article25} introduced a handwritten Hindi character classification using a combination of global character and local sub-unit features, resulting in a recognition accuracy of 93.5\%. Then, Chauhan et al. \cite{ref_article26} proposed a script-independent deep learning network evaluated across multiple scripts, including Hindi, establishing new benchmarks with performance improvements up to 11\%. Furthermore, a fusion-based hybrid-feature based on a combination of deep CNN features with handcrafted features has been explored, achieving a recognition accuracy of 98.7\% \cite{ref_article27}. The Integral Histogram of Oriented Displacement (IHOD) descriptor has also been used for handwritten Hindi script recognition, achieving high accuracy in character recognition and word spotting \cite{ref_article36}. This method outperformed several CNN-based models and contributed a large dataset of handwritten Hindi word images. These studies collectively contributed to the advancement of Hindi handwritten character recognition, offering diverse methodologies to address the complexities inherent in processing handwritten Indic scripts.This oversight draws attention towards the missed opportunity to fully harness the potential inherent within this domain of research. Therefore, a refined approach that meticulously accounts for the unique complexities of low-resource script is necessary. Equally, the availability of more consistent and standardized datasets is indispensable to facilitate model comparisons between various OCR models.
\section{Proposed Architecture}
\label{Architecture}
\begin{figure}[!t]
\centering
\includegraphics[width=0.5\textwidth]{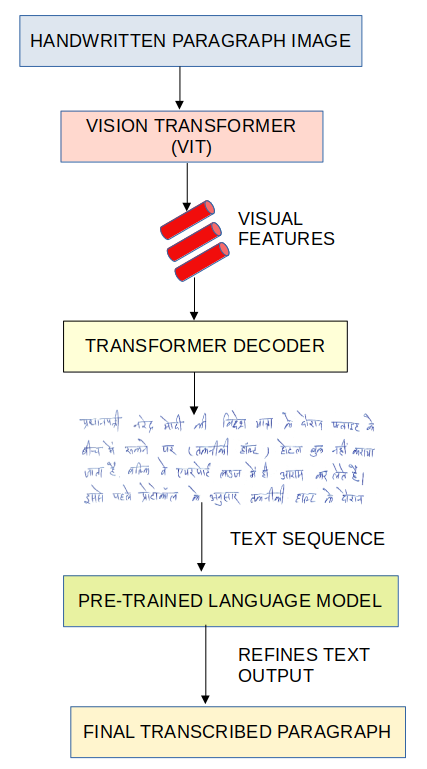}
\caption{The architecture consists of an image encoder, a transformer decoder, and a pre-trained language model. The image encoder extracts visual features from paragraph images, the transformer decoder generates an initial text sequence, and the language model refines it for coherence and accuracy.} 
\label{fig:1}
\end{figure}

The block diagram of the novel architecture proposed for paragraph-level handwritten text recognition in this paper is shown in Figure \ref{fig:1}. The proposed architecture begins by processing a handwritten paragraph image, which is divided into fixed-size patches and embedded into feature vectors. Positional encodings are also added to preserve spatial relationships, and a Vision Transformer (ViT) encoder extracts contextual visual features using multi-head self-attention. The extracted features are then flattened and projected into a sequence-compatible representation for the transformer decoder, which autoregressively generates text tokens using masked self-attention and cross-attention. The decoder aligns the generated tokens with ViT-extracted embeddings to ensure accurate transcription. A pre-trained language model (LM) further refines the generated text, enhancing fluency, coherence, and grammatical accuracy. The sequence is structured to maintain paragraph formatting and training is performed using cross-entropy loss combined with sequence-level optimization. Finally, the model provides a well-structured, refined transcription of the handwritten paragraph.

The proposed framework addresses the challenges of low-resource handwriting, such as varying cursive styles and inconsistent skews/slants, without relying on explicit segmentation. Moving away from traditional segmentation methods, the proposed model uses attention mechanisms to enhance text understanding and decoding, thereby improving efficiency and reducing recognition errors. 

\subsection{Pre-processing}
In our proposed framework, we performed pre-processing across all datasets to maintain model consistency and optimal input quality. We implemented a down-sampling technique using bi-linear interpolation to halve the resolution of images to 150 dpi, consistently applying this specification across datasets. The down-sampled images were then resized to 224x224 for pre-training and 448x448 for fine-tuning. This uniform approach to pre-processing ensured that the model was trained and evaluated under consistent conditions, enhancing its ability to generalize effectively across diverse datasets.
\subsection{Image Encoder}
The proposed network uses DeiTForMaskedImageModeling, an adaptation of the Data-efficient Image Transformer (DeiT) constructed for complex advanced masked image modeling challenges \cite{ref_article4}. This model is built on transformer architecture to extract intricate visual features from images.
\subsubsection{Knowledge Distillation}
\begin{figure}[!t]
\centering
\includegraphics[width=0.6\linewidth]{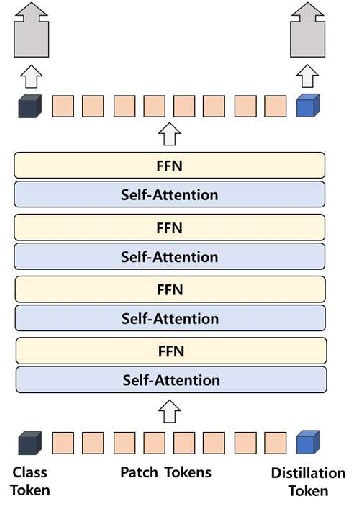}
\caption{Our distillation procedure by including a new distillation token. It interacts with the class and patch tokens through the self-attention layers. This distillation token is employed similarly to the class token, except that on the network output, its objective is to reproduce the (hard) label predicted by the teacher instead of a true label. Both the class and distillation tokens input to the transformers are learned by back-propagation.} 
\label{fig:2}
\end{figure}

Our proposed model uses a data-efficient image transformer as the image encoder. A knowledge distillation technique \cite{ref_article4}, as shown in Figure \ref{fig:2}, is used to pre-train this encoder. In knowledge distillation using a Vision Transformer \cite{ref_article3}, as the teacher model, the "distill token" emerges as a pivotal concept. This token is designed to capture and incorporate the distilled knowledge from the teacher model. During the training phase, it interacts with image patch embedding through the transformer's self-attention mechanism to closely mirror the teacher's output \cite{ref_article1}. The distill token's final state, refined through layers of attention, encapsulates the teacher's learned representations, thus serving as a compact vessel for transferring knowledge to the student model. This method enhances the student model's performance by imbuing it with the teacher's insights without necessitating the computational heft of the teacher model. The distill token illustrates an efficient strategy to use the complex patterns and relationships learned by larger models, facilitating the deployment of powerful yet lightweight models for various applications.

\subsubsection{Image Encoder Architecture}
 The image encoder architecture processes resized images of 224x224 pixels to ......., a choice that balances the need for detail retention against computational efficiency. The core of our framework is built upon a deep transformer structure with 12 hidden layers and a hidden size of 768, designed to process complex visual data. It incorporates 12 attention heads to focus on multiple image features, enhancing feature extraction. The model processes the input images by breaking them into 16x16 pixel patches, treating them as sequences, thereby combining the transformer’s sequential processing capabilities with the spatial characteristics of images for more detailed analysis. We use the Gaussian Error Linear Unit (GELU) as the activation function and layer normalization to ensure stable learning dynamics. The model’s intermediate layer size of 3072 allows complex transformations in a higher-dimensional space. The encoder stride of 16 ensures efficient image processing. This DeiTForMaskedImageModeling model exemplifies a sophisticated application of transformers in computer vision, optimizing both performance and data efficiency.

\subsection{Attention Mechanism}
Attention mechanisms enable models to selectively focus on relevant parts of the input data, enhancing decision-making processes. Originally developed for natural language processing (NLP), these mechanisms help models dynamically prioritize different elements, such as words in a sentence. The concept has been expanded to applications in vision and audio processing. Cross-modal attention \cite{ref_article5} is a further evolution of these mechanisms that allows models to integrate and focus on information from different data types, such as text and images. This invention encourages a more comprehensive understanding through the smooth integration of textual and visual features. It is especially valuable in tasks necessitating the synthesis of multi-modal inputs for predictions.

\subsection{Language Model}
\begin{figure}[ht]
    \subfigure{\includegraphics[width=0.42\columnwidth]{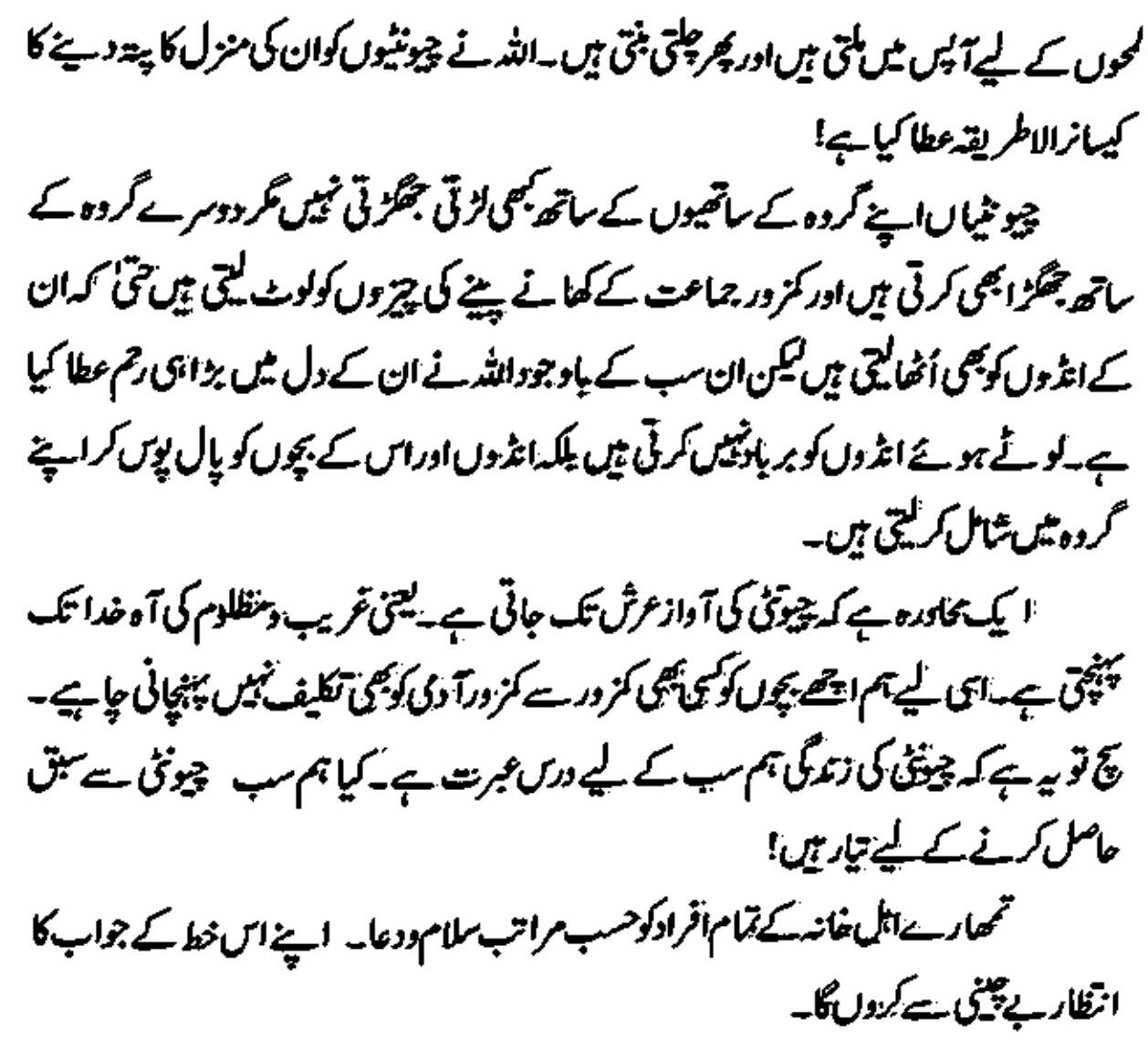} }
    \hfill
    \subfigure{\includegraphics[width=0.45\columnwidth]{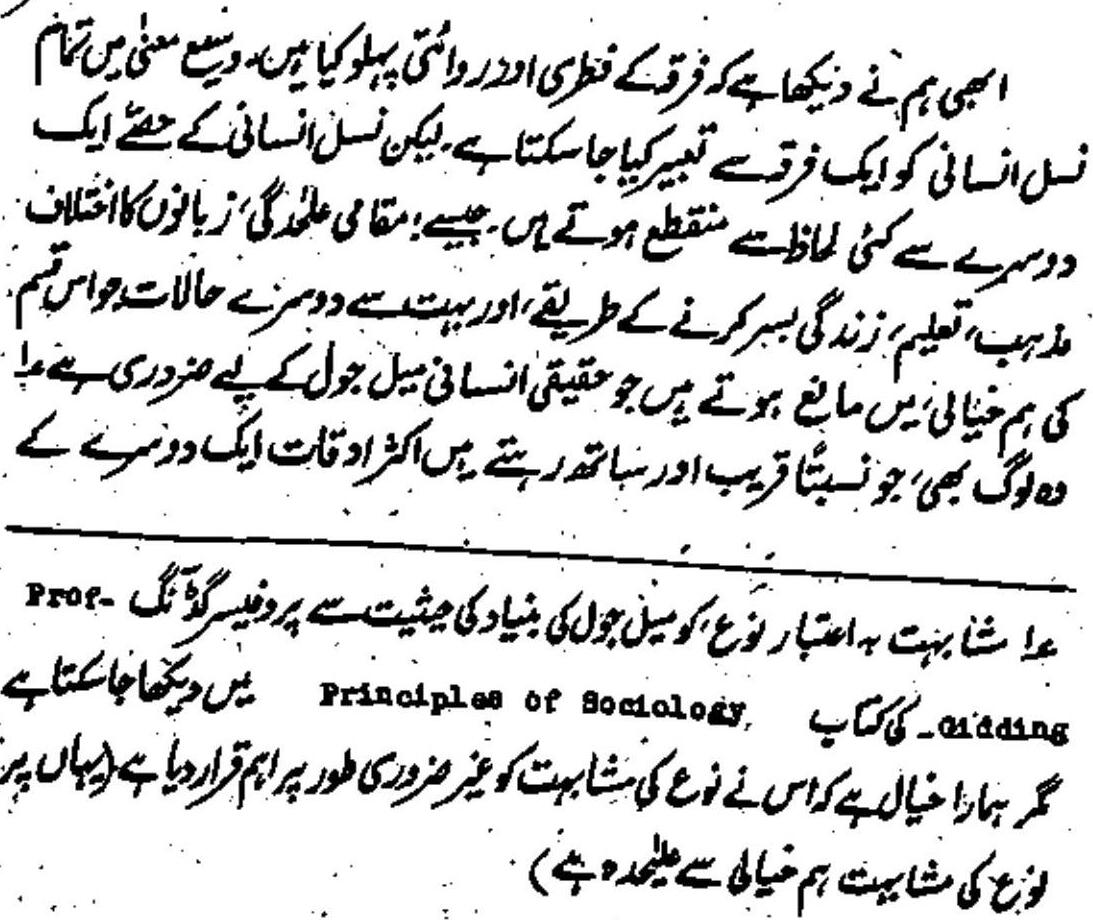} }
    \caption{Printed Urdu Text Paragraphs from different books of science and philosophy}
    \label{fig:4}
\end{figure}

In developing our proposed framework, we adopt a language transformer-based architecture \cite{ref_article6} to improve the text sequence obtained from our proposed transformer decoder model. The language model was pre-trained for masked language modeling task, as shown in Figure \ref{fig:4}, to enhance its linguistic capabilities. The model contains six hidden layers, and each layer is designed with a hidden size of 768 units and 12 attention heads. Each layer has multiple attention heads facilitating attention mechanisms across various input data segments. This configuration is tailored to process sequences with a maximum length of 512 tokens, supported by absolute position embedding to retain the syntactic and semantic order of language input. The architecture incorporates a 10 \% dropout rate in the hidden layers and attention probabilities to avoid overfitting and generalizing the model. The model's activation function of choice, GELU, ensures smooth non-linearity and the ability to model complex relationships in the data. Our model benefits from a vocabulary size of 50,026 tokens to learn the intricacies of the Urdu language. This architectural design emphasizes our model's capability to learn from masked linguistic contexts and predictively model language with remarkable accuracy and speed.
\subsection{Decoder}
The Transformer decoder, following a autoregressive structure \cite{ref_article29}, converts visual embeddings extracted by the Vision Transformer (ViT) into textual tokens for paragraph-level handwritten text recognition. The decoder comprises multi-head self-attention, cross-attention for alignment with ViT features, and feedforward layers to model long-range dependencies. During training, teacher forcing is employed, where the ground-truth token at each time step is provided as input for the next step instead of the previously generated token, ensuring stable convergence. The decoder is trained on paired image-text datasets using cross-entropy loss to minimize discrepancies between predicted and ground-truth sequences. To mitigate exposure bias, scheduled sampling is incorporated to gradually reduce reliance on teacher forcing. During inference, the model generates text autoregressively, predicting one token at a time while using its own outputs as subsequent inputs. Decoding strategies, such as beam search, nucleus sampling, and greedy decoding, enhance the quality of text generation. This approach ensures accurate, coherent, and contextually structured paragraph-level text recognition by leveraging self-attention, cross-attention, and language modeling techniques.
\begin{figure}[h]
\centering
\includegraphics[width=0.85\textwidth]{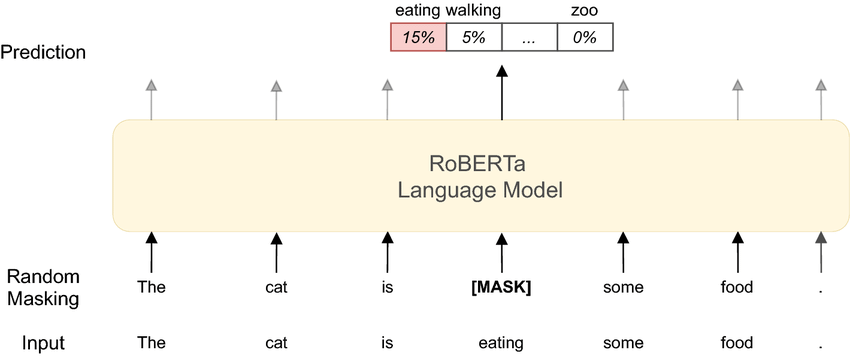}
\caption{RoBERTa—masked language modeling with the input sentence: The cat is eating some food}
\label{fig:3}
\end{figure}
\subsection{Fine-tuning The Network}
The whole network, which consists of the image encoder, the transformer decoder,the language model, is fine-tuned using the Parimal Urdu and Hindi datasets, which consists of paragraph images. The whole architecture, when fine-tuning, is shown in Figure \ref{fig:1}. During fine-tuning, the Image encoder processes a paragraph image and outputs visual features based on the visual tokens learned during pre-training. The transformer decoder outputs a sequence of text. The pre-trained language model then refines the output text contextually based on learned tokens and provides the final text output.
\section{Experimental Analysis and Discussion}
\label{Experimental Analysis and Discussion}
\subsection{Datasets}
\begin{figure}[ht]
    \subfigure{\includegraphics[width=0.42\columnwidth]{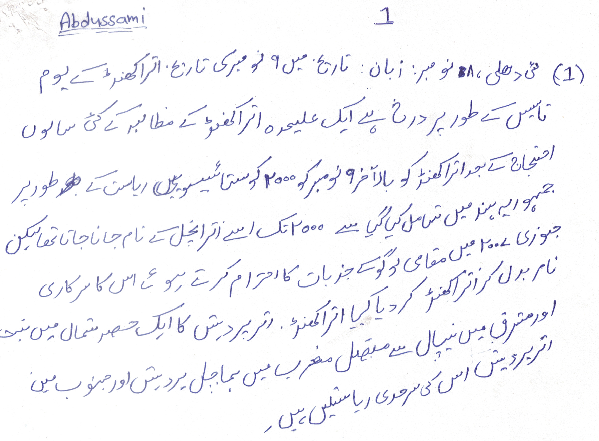}}
    \hfill
    \subfigure{\includegraphics[width=0.42\columnwidth]{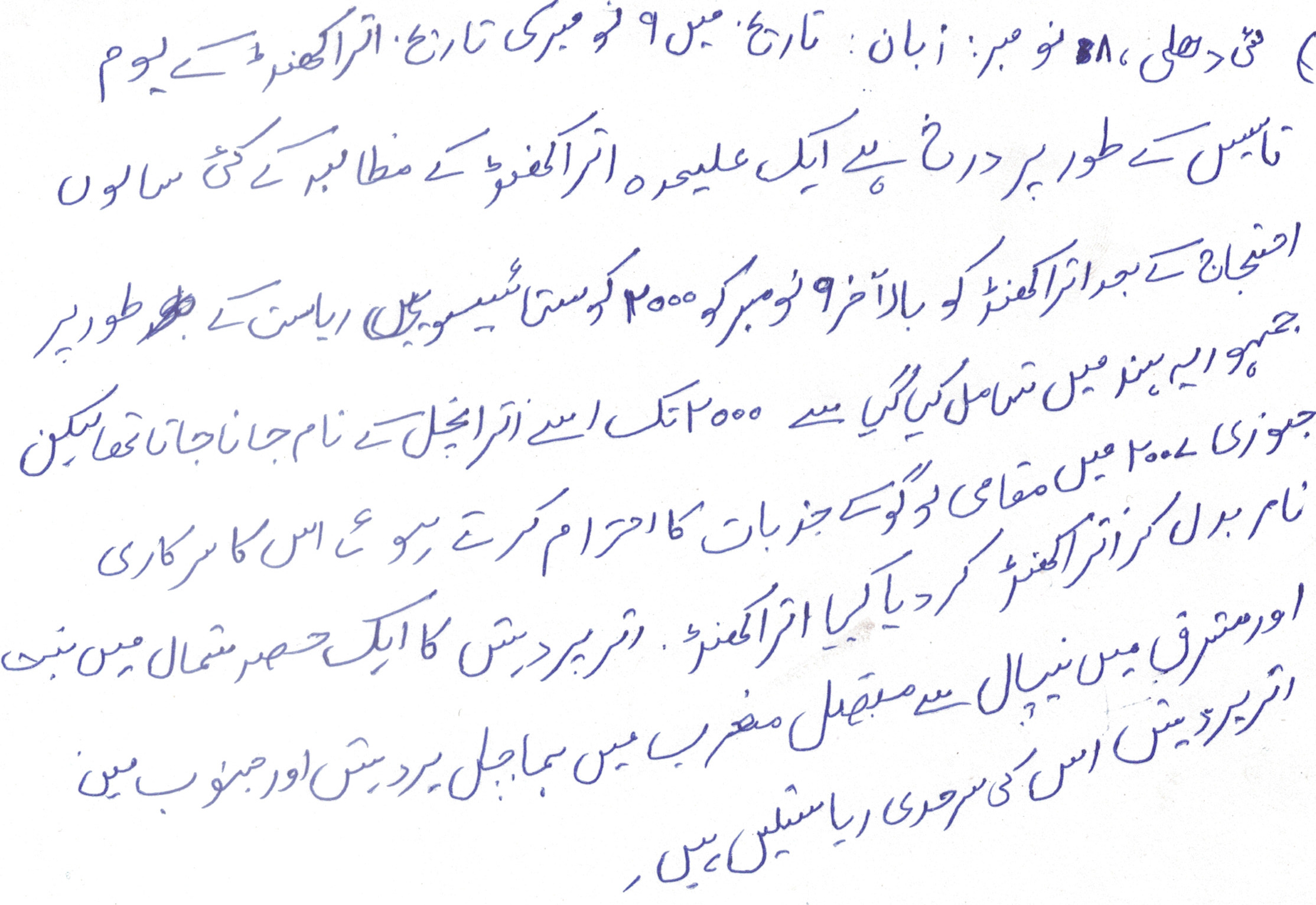}}
    
    \caption{Handwritten text paragraphs from two authors with quite different writing styles from the Parimal Urdu dataset.}
    \label{fig:5}
\end{figure}

We used various datasets to pre-train our image encoder and language model, fine-tune them on handwritten documents and finally test the proposed framework.  
\paragraph{Printed Datasets:}
To pre-train our DeiT model, we created a dataset of 21,000 images, as shown in Figure \ref{fig:5}, from various online sources, focusing on the diversity of Urdu printed text. This dataset includes multiple font styles and layouts to comprehensively represent printed Urdu and Hindi text, which helps the model learn the intricacies of Urdu text recognition. The diversity of data in the training phase is crucial for helping the model generalize across different textual formats and provide a strong foundation for further fine-tuning for specialized tasks, ensuring robust initial training for enhanced performance in advanced applications.

\begin{figure}[ht]
    \subfigure{\includegraphics[width=0.5\columnwidth]{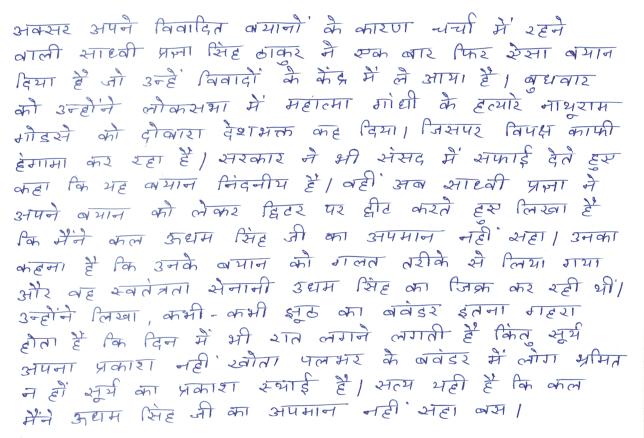}}
    \hfill
    \subfigure{\includegraphics[width=0.5\columnwidth]{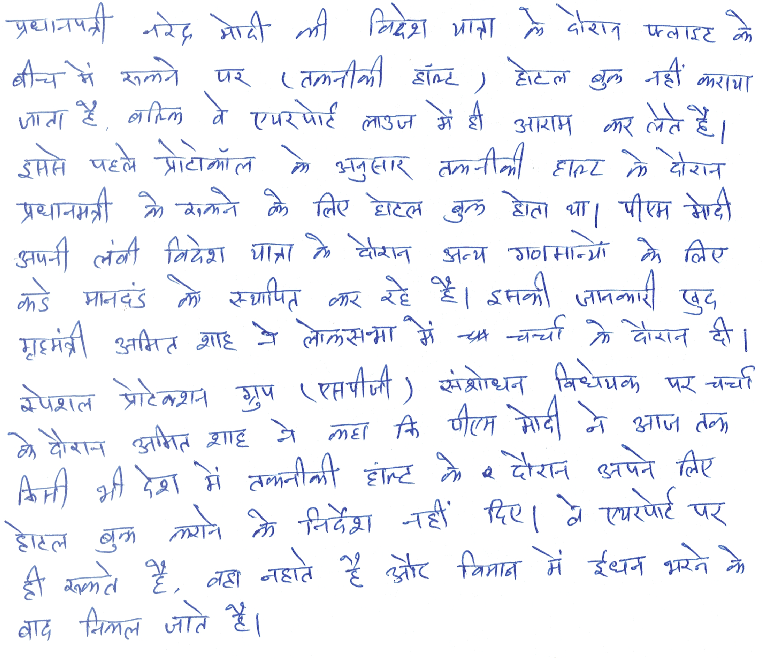}}
    
    \caption{Handwritten text paragraphs from two authors with quite different writing styles from the Parimal Hindi dataset.}
    \label{fig:6}
\end{figure}

\paragraph{Handwritten Paragraph Dataset (Parimal Urdu and Parimal Hindi):}
We also used the Parimal Urdu and Parimal Hindi dataset, which consists of 500 pages of handwritten Urdu and Hindi text as shown in Figure \ref{fig:6}. It was contributed by 10 individuals from various age groups, each selected for their different handwriting styles shown in Figure \ref{fig:6}. This variety introduces 10 unique styles into the dataset, providing a solid basis for evaluating the adaptability and effectiveness of our model with various natural writing variations. This dataset was crucial for the fine-tuning and evaluation phases.

\paragraph{Text Data:}
In our research, the RoBERTa model was subjected to an extensive pre-training phase using a carefully curated corpus of Urdu text. This corpus, rich in diversity, consists of 20 million lines of text, amounting to approximately 200 million words, drawn from various sources, including news articles and magazines. This selection aimed to capture the linguistic richness and variability of the Urdu script. The sheer volume and variety of the dataset provided a solid foundation for the model, enabling it to acquire a deep understanding of Urdu syntax, semantics, and contextual nuances. Such comprehensive pre-training is crucial, as it significantly enhances the model's performance across various NLP tasks. 

\paragraph{Public Handwritten Text Lines Dataset:}
\textbf{PUCIT-OHUL} \cite{ref_article23} dataset consists of multiple writing styles, different types of pen, ink types, text sizes and background types and colors. It was collected from 100 students between 20 and 24 years of age. A total of 479 pages of text were collected. Each page was scanned at 200 DPI and text lines were manually segmented. The dataset contains a total of 7401 text lines and 80,059 words.
\textbf{NUST-UHWR} \cite{ref_article18} dataset is obtained from various websites, including social networks and news websites, and contains 10,606 samples of handwritten Urdu text lines.
\subsection{Data Augmentation}
To improve accuracies obtained from our proposed model and prevent overfitting, we developed a comprehensive data augmentation strategy, which was implemented only during the training phase. This strategy includes a variety of techniques, such as resolution modification, perspective changes, elastic distortion, and adjustments in brightness and contrast, each with a 0.2 probability of use. This systematic and diversified augmentation approach significantly enhances the model's ability to generalize features and adapt to real-world data variations, which is crucial for robust pattern recognition performance.
\subsection{Metrics}
To evaluate the performance of our UHHTR model, we considered three commonly used metrics, Character Error Rate (CER), Word Error Rate (WER), and Line Error Rate (LER), which indicate the accuracy of the proposed model in recognizing individual characters, whole words, and Lines, respectively. These metrics are calculated using the Levenshtein distance (denoted by $\text{levd}$), which measures the difference between the ground truth ($\mathbf{y}$) and the recognized text ($\hat{\mathbf{y}}$), normalized by the total length of the ground truth ($\text{len}(\mathbf{y})$). This normalization process ensures that discrepancies in the shortest lines are proportionately weighted against those in longer segments, providing a fair metric across texts of varying lengths. The character, word and line recognition rates were calculated by subtracting their respective error rates from 100. The following equations were used to calculate CER, WER and LER.
\begin{equation}
\text{CER} = \frac{\sum_{i=1}^{K} \text{len}_i^y}{\sum_{i=1}^{K} \text{lev}_i^d(\hat{y}_i, y_i)}
\label{eq:7}
\end{equation}
\begin{equation}
\text{WER} = \frac{\sum_{i=1}^{K} \text{word count in } y_i}{\sum_{i=1}^{K} \text{lev}_i^d(\text{words in } \hat{y}_i, \text{words in } y_i)}
\label{eq:8}
\end{equation}
Here, K represents the total number of images in the dataset. For WER, punctuation characters are treated as independent words aligned with conventions.
\begin{equation}
\text{LER} = \frac{1}{N} \sum_{i=1}^{N} \text{lev}^d(\hat{y}_i, y_i)
\label{eq:9}
\end{equation}
Where $N$ is the total number of sentences in the dataset and $\text{lev}(\hat{y}_i, y_i)$ is the Levenshtein distance between the predicted line $\hat{y}_i$ and the ground truth line $y_i$ for the $i$-th sentence. The denominator, $\sum_{i=1}^{N} 1$, counts the total number of sentences in the ground truth, providing a normalization factor for the error rate.

We also employed two metrics to assess our model's text segmentation accuracy: Intersection over Union (IoU) and mean Average Precision (mAP). IoU measures the overlap of pixels classified as text versus the union of such pixels compared to the ground truth, normalized across images. mAP calculates average precision across IoU thresholds from 50\% to 95\% in 5\% increments, weighted by pixel count for a dataset-wide score. 

\subsection{Experimental results}
We first used the Parimal Urdu dataset comprising 500 images of handwritten Urdu and Hindi paragraphs for experimentation. The dataset was methodically partitioned into training, testing, and validation subsets, with respective proportions of 70\%, 20\%, and 10\%. This split ensures a balanced approach, allowing for comprehensive training while retaining sufficient data to test and validate the models. The results were evaluated based on three key metrics, CRR, WRR and LRR. The results of these experiments are presented in Table~\ref{tab:1} and Table~\ref{tab:2}.

\begin{table}[!t]
    \centering
    \caption{Recognition Rates (\%) of the Proposed Model on Paragraph Urdu Data}
    \label{tab:1}
    \renewcommand{\arraystretch}{1.2}
    \begin{tabular}{|l|c|c|c|c|c|c|}
        \hline
        \multirow{2}{*}{Models} & \multicolumn{2}{c|}{CRR} & \multicolumn{2}{c|}{WRR} & \multicolumn{2}{c|}{LRR} \\
        \cline{2-7}
        & Val & Test & Val & Test & Val & Test \\
        \hline
        Proposed HWR & 94.80 & 95.20 & 83.60 & 84.70 & 72.78 & 73.24 \\
        \hline
    \end{tabular}
\end{table}
\begin{table}[!t]
    \centering
    \caption{Recognition Rates (\%) of the Proposed Model on Paragraph Hindi Data}
    \label{tab:2}
    \renewcommand{\arraystretch}{1.2}
    \begin{tabular}{|l|c|c|c|c|c|c|}
        \hline
        \multirow{2}{*}{Models} & \multicolumn{2}{c|}{CRR} & \multicolumn{2}{c|}{WRR} & \multicolumn{2}{c|}{LRR} \\
        \cline{2-7}
        & Val & Test & Val & Test & Val & Test \\
        \hline
        Proposed HWR & 80.64 & 78.20 & 70.60 & 67.65 & 54.78 & 57.24 \\
        \hline
    \end{tabular}
\end{table}
\subsection{Comparative Study}
To compare the performance of our proposed model with state-of-the-art models we considered nine different methods from the literature. The experiments performed on two public datasets for Urdu text recognition, the NUST-UHWR and PUCIT-OHUL datasets, and the results were calculated basedon CRR and WRR. The results obtained from the prior models and our proposed model On the NUST-UHWR dataset is shown in Table \ref{tab:3}. From Table \ref{tab:3}, it is clear that our proposed model performs better and sets a new benchmark. Table \ref{tab:4} shows a comparison study on the PUCIT-OHUL dataset with existing models. The results indicate that our proposed model outperforms existing models in both CRR and WRR. In the case of character recognition, the proposed model is superior to any existing models, showing that visual and textual tokens are mapped appropriately. The existing models based on CNNs do not perform well on Urdu text because they cannot identify the intricacies of Urdu handwritten text recognition. The models based on CNNs and transformers also suffer as they cannot process the contextual information in the text. It is important noting that as the proposed approach uses a language model to derive contextual information, and a multi-modal attention network to extract visual details, it improved handwritten Urdu text recognition. In addition, the proposed model suffers in word recognition compared to character recognition due to the use of inconsistent spacing, which is quite natural in Urdu scripts.  

\begin{table}[!t]
    \centering
    \caption{Comparison of different models for offline Urdu handwritten text recognition on NUST-UHWR Dataset}
    \label{tab:3}
    
    \begin{tabular}{|l|c|c|}
        \hline
        Models & CRR\% & WRR\% \\
        \hline
        BLSTM [12] & 72.60 & 61.37\\
        Modified CRNN [13] & 81.50 & 70.60 \\
        MDLSTM [14] & 85.87 & 74.60 \\
        CNN-RNN [15] & 86.75 & 76.05 \\
        BiGRU [16] & 86.30 & 75.02 \\
        TrOCR [17] & 79.88 & 68.53 \\
        Conv. Recursive [18] & 92.75 & 81.88 \\
        Conv. Transformer [19] & 94.03 & 83.13 \\
        Unified Arch. [20]& 94.10 & 83.24 \\
        \hline
        \textbf{Proposed} & \textbf{96.24} & \textbf{85.3} \\
        \hline
    \end{tabular}
\end{table}

\begin{table}[!t]
    \centering
    \caption{Comparison of different models for offline Urdu handwritten text recognition on PUCIT-OHUL dataset}
    \label{tab:4}
    \begin{tabular}{|l|c|c|}
        \hline
        Models & CRR\% & WRR\% \\
        \hline
        CNN-GRU & 33.00 & 22.44 \\
        CNN-LSTM & 32.04 & 21.42 \\
        CNN-BGRU & 32.76 & 23.16 \\
        CNN-BLSTM [7] & 32.02 & 22.23 \\
        SimpleHTR [9] & 14.94 & 5.09 \\
        LineHTR [8] & 30.05 & 18.19 \\
        CRNN [10] & 32.86 & 24.18 \\
        CALText[11] & 82.06 & 51.97 \\
        \hline
        \textbf{Proposed} & \textbf{92.05} & \textbf{80.54}\\
        \hline
    \end{tabular}

\end{table}

\section{Conclusion and Future Scope}
\label{Conclusion and Future Scope}

In segmentation-based approaches, the recognition of characters, words, and lines was severely affected due to wrong manual segmentation, as the Urdu script follows a cursive writing style. This study introduces a novel segmentation-free approach to recognizing handwritten Urdu datasets using multi-modal attention. This method is further refined by fine-tuning with targeted datasets for handwritten Urdu recognition. In addition, a self-collected dataset, Parimal Urdu, is introduced in this research work. The proposed framework is evaluated on three different datasets, including Parimal, PUCIT-OHUL and NUST-UHWR. Notably, the new model achieved recognition accuracies that were higher than those of previous methods. In fact, it achieves state-of-the-art results on these datasets. Its implicit line segmentation process enables the recognition of inclined lines and paragraphs. For future work, this model can be extended to be applied to more diverse text recognition tasks, such as Patwari Urdu, Arabic, and Persian text recognition.

\end{document}